\documentclass[pdflatex,sn-basic]{sn-jnl}

\usepackage{graphicx}%
\usepackage{multirow}%
\usepackage{amsmath,amssymb,amsfonts}%
\usepackage{amsthm}%
\usepackage{mathrsfs}%
\usepackage[title]{appendix}%
\usepackage{xcolor}%
\usepackage{textcomp}%
\usepackage{manyfoot}%
\usepackage{booktabs}%
\usepackage{algorithm}%
\usepackage{algorithmicx}%
\usepackage{algpseudocode}%
\usepackage{listings}%
\usepackage[figuresright]{rotating}
\usepackage{comment}
\usepackage{hyperref}
\usepackage{url}
\usepackage{csquotes}
\usepackage{setspace}
\usepackage{forest}    
\usetikzlibrary{positioning,arrows.meta} 
\usepackage{adjustbox} 
\usepackage{pifont}

\begin{document}

\title{Towards a Data-Centric Understanding of Spatial Reasoning in Multimodal Large Language Models}

\author*[1]{\fnm{Mahdieh} \sur{Mallahnezhad}}\email{mallahne@ualberta.ca}
\author[1]{\fnm{Masoud} \sur{Jafaripour}}\email{jafaripo@ualberta.ca}
\author[1]{\fnm{Mohamadali} \sur{Tofigh}}\email{tofigh@ualberta.ca}
\author[1]{\fnm{Sheila} \sur{Schoepp}}\email{sschoepp@ualberta.ca}
\author[1,2]{\fnm{Osmar R.} \sur{Zaiane}}\email{zaiane@ualberta.ca}

\affil[1]{%
    \orgdiv{Department of Computing Science},
    \orgname{University of Alberta},
    \orgaddress{%
        \city{Edmonton},
        \state{Alberta},
        \country{Canada}}}

\affil[2]{%
    Alberta Machine Intelligence Institute (Amii) Fellow,
    CIFAR Fellow}

\abstract{Multimodal Large Language Models (MLLMs) have been widely applied in embodied agents and robotics for vision tasks. However, MLLMs suffer from limitations in spatial reasoning and perception. Post-training techniques have shown promise for improving these capabilities; yet, there are limited data-driven surveys of such approaches. In this survey, we consider approaches to fine-tuning pretrained MLLMs for spatial reasoning tasks, ranging from supervised to reinforcement learning, with an emphasis on the datasets used. We investigate themes of dataset construction, data modality and annotation, and evaluation of the resulting models. Based on these themes, we categorize research into three groups: dataset design, dataset construction and evaluation. We find four main methods for constructing fixed spatial annotations: rules and templates, LLM prompting, procedural generation, and human-in-the-loop. We also observe substantial variation in evaluation approaches, including different scoring methods, tolerance metrics, and benchmark types. Finally, we identify opportunities for future research in unified dynamic annotation, efficient spatial reasoning, standardization of evaluation, expanded spatial range, and human-centered spatial interactions.}


\maketitle

\section{Introduction}
\label{sec:introduction}

Vision-Language Models (VLMs) and, more broadly, Multimodal Large Language Models (MLLMs) combine visual and language reasoning to interpret visual inputs, understand spatial and semantic relations between objects in those inputs, and produce answers to natural language questions based on them~\citep{Liu2023VisualIT}.
Although these models perform well on image captioning, visual question answering, and object identification~\citep{yin2024survey}, they struggle with spatial reasoning~\citep{Cheng2024SpatialRGPTGS}. For example, they often fail when asked questions about spatial relations that should be relatively easy to answer, such as relative positions (\enquote{Is the cup to the left of the plate?}), distances (\enquote{How far is the table from the wall?}), size comparisons (\enquote{Which object is bigger?}), and even different view angles (\enquote{What would this scene look like from behind?})~\citep{Chen2024SpatialVLMEV, Huang2025SpatialDISEAU}. 
This presents a significant problem: even though MLLMs can analyze and describe objects within a scene, they can make incorrect inferences about those objects' locations or their spatial relationships with each other. 

Spatial understanding is the ability to make inferences about spatial layouts and spatially relevant properties of objects. Humans use it in everyday activities: to navigate around a crowded room, to place an object on a shelf, or to decide whether a box fits in a particular location~\citep{Vieu1997SpatialRA}. It is not just a perceptual ability; spatial understanding plays an essential role in communicative interactions, action planning, and reasoning about the physical world~\citep{Tversky2005FunctionalSO}. 

Spatial reasoning has accordingly become a central challenge for embodied artificial intelligence (AI)~\citep{Liu2025SpatialII}. 
An embodied agent must be able to perceive, plan, and act in the physical world. Without a firm grasp of spatial relations, agents will struggle to map perceptions to the corresponding actions~\citep{Yuan2024RoboPointAV}. 
Spatial understanding therefore acts as a bottleneck on the deployment of embodied VLM/MLLM agents in real-world environments~\citep{Song2025RoboSpatialTS}. 

A number of mechanisms allow MLLMs to complete tasks involving spatial reasoning. The model architecture determines how visual and linguistic information are represented; e.g., how RGB data are combined with depth data can have a significant effect on a model’s spatial reasoning abilities~\citep{Cai2025SpatialBotPS,Liu2025SSRED}. The training method also affects spatial reasoning outcomes; e.g., post-training with supervised fine-tuning (SFT) and reinforcement learning (RL) using verifiable rewards can have implications for how the models produce outcomes~\citep{Ouyang2025SpaceRRM,Batra2025SpatialThinkerR3}. 

Another perspective suggests that it is the training corpus that is a key factor limiting MLLMs’ spatial reasoning abilities. In~\citet{Chen2024SpatialVLMEV}, the authors hypothesize that the limitations in current VLMs are not due to the model architecture itself, but due to the lack of spatial information present in the training data. This hypothesis is consistent with~\citet{Ogezi2025SpaREES}, who found that spatial relations are underrepresented in the datasets they examined and that the performance of VLMs can be improved by introducing synthetic spatial relations. While these studies do support the importance of spatial supervision, they do not eliminate the potential influence of architecture, training objective, or the general-purpose nature of the training data.
These results encourage research that focuses on the data used for training and the creation of datasets for spatial reasoning, employing strategies that range from procedurally-generated synthetic scenes to large-scale automatic annotation of real-world images~\citep{Deng2025InternSpatialAC,Cheng2024SpatialRGPTGS,Song2025RoboSpatialTS}. 

Recent surveys of MLLMs have explored the area of spatial reasoning from either an architecture- and training-focused~\citep{Zheng2025MultimodalSR} or task and benchmark perspective ~\citep{Liu2025SpatialRI}. There are also data-centric surveys that explore the creation and utilization of multimodal training datasets\citep{Bai2024ASO}.
There are also broader surveys that look at data-driven methods and spatially trained datasets (see Section~\ref{relatedSurvey}). Our survey adds to these by looking at dataset creation pipelines, annotation sources, and comparison of evaluation methods as our primary focus. Since the data sources and their design appear to be one of the major factors limiting current models’ performance on these tasks~\citep{Chen2024SpatialVLMEV,Deng2025InternSpatialAC}, such a survey can help guide future research in this area. We do so by reviewing datasets used to fine-tune pre-trained foundation models for spatial reasoning and the evaluation of the models produced. We use the term fine-tuning broadly to refer to any update to model parameters/structure after pretraining, including supervised instruction tuning, parameter-efficient tuning (e.g., PEFT/LoRA), preference tuning, and RL with spatially grounded rewards ~\citep{Li2025SpatialLadderPT,Shen2025FineGrainedPO,Wu2025ReinforcingSR}.\

We focus on selected training datasets and pipelines that produced them, and we discuss benchmark datasets only when they accompany a training dataset, in order to contrast learning- versus evaluation-focused design choices. Table~\ref{tab:spatial_reasoning} summarizes the datasets we review. In short, the contributions of this survey are:\

\begin{figure}[!t]
    \centering
    \begin{adjustbox}{max width=\linewidth, max totalheight=0.9\textheight}
\definecolor{taxroot}{HTML}{73A9C2}
\definecolor{taxgrnD}{HTML}{80ae9a}
\definecolor{taxgrnM}{HTML}{A0D6B4}
\definecolor{taxgrnL}{HTML}{F5FFFA}
\definecolor{taxbluD}{HTML}{D4AF37}
\definecolor{taxbluM}{HTML}{EEDC82}
\definecolor{taxbluL}{HTML}{FFFFED}
\definecolor{taxperD}{HTML}{E4717A}
\definecolor{taxperM}{HTML}{FFB7C5}
\definecolor{taxperL}{HTML}{FFF0F5}


\providecommand{\taxcite}{}%
\renewcommand{\taxcite}[1]{\textcolor{black!60}{\citep{#1}}}%

\ifdefined\taxdesignw\else\newlength{\taxdesignw}\fi
\ifdefined\taxcolw\else\newlength{\taxcolw}\fi
\ifdefined\taxinw\else\newlength{\taxinw}\fi
\setlength{\taxdesignw}{1.4\textwidth}%
\setlength{\taxcolw}{\dimexpr(\taxdesignw-20pt)/3\relax}%
\setlength{\taxinw}{\dimexpr\taxcolw-8pt\relax}%
\begin{tikzpicture}[
  font=\scriptsize,
  every node/.style={line width=0.3pt},
  hdr/.style={draw=black!65, rounded corners=2.5pt, align=center,
              font=\scriptsize\bfseries, inner ysep=4pt,
              text width=\dimexpr\taxcolw-10pt\relax},
  cat/.style={draw=black!60, rounded corners=2.5pt, align=center,
              font=\scriptsize\bfseries, inner ysep=3.5pt,
              text width=\dimexpr\taxinw-10pt\relax},
  leaf/.style={draw=black!55, rounded corners=2pt, align=left,
               inner xsep=4pt, inner ysep=2.6pt,
               text width=\dimexpr\taxinw-10pt\relax},
  spine/.style={draw=black!60, line width=0.3pt},
  rootarrow/.style={draw=black!70, line width=0.4pt, -{Stealth[length=4.5pt]}},
]
\coordinate (colA) at (0,0);
\coordinate (colB) at (\dimexpr\taxcolw+10pt\relax,0);
\coordinate (colC) at (\dimexpr2\taxcolw+20pt\relax,0);
\node[hdr, fill=taxgrnD, anchor=north west] (hA) at (colA) {Dataset Design Dimensions};
\node[hdr, fill=taxbluD, anchor=north west] (hB) at (colB) {Dataset Building};
\node[hdr, fill=taxperD, anchor=north west] (hC) at (colC) {Evaluation};
\node[hdr, fill=taxroot, text width=0.45\taxdesignw, anchor=south, font=\small\bfseries]
  (root) at ([yshift=26pt]hB.north)
  {Data-Centric Spatial Reasoning in MLLMs};
\draw[rootarrow] (root.south) -- +(0,-7pt) -| (hA.north);
\draw[rootarrow] (root.south) -- (hB.north);
\draw[rootarrow] (root.south) -- +(0,-7pt) -| (hC.north);
\node[cat, fill=taxgrnM, anchor=north west] (Ac1) at ([xshift=8pt,yshift=-7pt]hA.south west) {Ground Truth};
\node[leaf, fill=taxgrnL, below=2.6pt of Ac1] (Al1) {Human supervision pipeline \taxcite{Chen2025Robo2VLMVQ,Huang2025SpatialDISEAU}};
\node[leaf, fill=taxgrnL, below=2.6pt of Al1] (Al2) {Autonomous generation pipeline \taxcite{Chen2024SpatialVLMEV,Deng2025InternSpatialAC}};
\node[cat, fill=taxgrnM, below=7pt of Al2] (Ac2) {Patterns of Tasks};
\node[leaf, fill=taxgrnL, below=2.6pt of Ac2] (Al3) {Perceptual reasoning (e.g., identifying an object) \taxcite{Li2025SpatialLadderPT}};
\node[leaf, fill=taxgrnL, below=2.6pt of Al3] (Al4) {Relational reasoning (e.g., left-of, larger than) \taxcite{Cheng2024SpatialRGPTGS}};
\node[leaf, fill=taxgrnL, below=2.6pt of Al4] (Al5) {Temporal reasoning (e.g., ``into'' or ``onto'') \taxcite{Ouyang2025SpaceRRM,Wang2025VideoSTRRM}};
\node[leaf, fill=taxgrnL, below=2.6pt of Al5] (Al6) {Metric reasoning (e.g., estimating object size) \taxcite{Cai2025SpatialBotPS}};
\node[leaf, fill=taxgrnL, below=2.6pt of Al6] (Al7) {Occlusion-aware reasoning (e.g., targets not visible in the query view) \taxcite{Lee2025SpatialMosaicAM}};
\node[leaf, fill=taxgrnL, below=2.6pt of Al7] (Al8) {Planning (e.g., action sequences to achieve spatial objectives) \taxcite{Sun2025SpaceVistaAV}};
\node[leaf, fill=taxgrnL, below=2.6pt of Al8] (Al9) {Perspective-transformation reasoning (e.g., matching objects across different camera poses) \taxcite{Deng2025InternSpatialAC}};
\node[cat, fill=taxgrnM, below=7pt of Al9] (Ac3) {Spatial Viewpoints};
\node[leaf, fill=taxgrnL, below=2.6pt of Ac3] (Al10) {Single-view: inferring a mug behind a laptop from one 2D scene \taxcite{Chen2024SpatialVLMEV}};
\node[leaf, fill=taxgrnL, below=2.6pt of Al10] (Al11) {Multi-view: implicit mental model from multiple observations \taxcite{Deng2025InternSpatialAC}};
\node[cat, fill=taxgrnM, below=7pt of Al11] (Ac4) {Frames of Reference};
\node[leaf, fill=taxgrnL, below=2.6pt of Ac4] (Al12) {Egocentric frame (e.g., ``to my left'' from the observer's view) \taxcite{Zhao2025EmbodiedRCF}};
\node[leaf, fill=taxgrnL, below=2.6pt of Al12] (Al13) {Object-centric frame (e.g., a bowl is ``in front of'' a car) \taxcite{Song2025RoboSpatialTS}};
\node[leaf, fill=taxgrnL, below=2.6pt of Al13] (Al14) {World-centric frame (e.g., something being ``north of'' another) \taxcite{Chen2024SpatialVLMEV}};
\node[leaf, fill=taxgrnL, below=2.6pt of Al14] (Al15) {Top-down representation (e.g., a flat 2D map view) \taxcite{Ouyang2025SpaceRRM}};
\draw[spine] ([xshift=3pt]hA.south west) |- (Ac1.west);
\draw[spine] ([xshift=3pt]hA.south west) |- (Ac2.west);
\draw[spine] ([xshift=3pt]hA.south west) |- (Ac3.west);
\draw[spine] ([xshift=3pt]hA.south west) |- (Ac4.west);
\node[cat, fill=taxbluM, anchor=north west] (Bc1) at ([xshift=8pt,yshift=-7pt]hB.south west) {Construction Strategies};
\node[leaf, fill=taxbluL, below=2.6pt of Bc1] (Bl1) {Ground-up: build the dataset corpus from scratch \taxcite{Tang2025SparkleMB,Huang2025SpatialDISEAU}};
\node[leaf, fill=taxbluL, below=2.6pt of Bl1] (Bl2) {Reformatting: turn existing vision datasets into spatial QA pairs \taxcite{Ranasinghe2024LearningTL,Song2025RoboSpatialTS}};
\node[leaf, fill=taxbluL, below=2.6pt of Bl2] (Bl3) {Mixing sources: combine indoor scans, outdoor scenes, and aerial videos \taxcite{Sun2025SpaceVistaAV}};
\node[cat, fill=taxbluM, below=7pt of Bl3] (Bc2) {Annotation Approaches};
\node[leaf, fill=taxbluL, below=2.6pt of Bc2] (Bl4) {Template- and rule-based \taxcite{Chen2024SpatialVLMEV,Cheng2024SpatialRGPTGS}};
\node[leaf, fill=taxbluL, below=2.6pt of Bl4] (Bl5) {LLM-based \taxcite{Ogezi2025SpaREES,Ning2025EnhancingSR}};
\node[leaf, fill=taxbluL, below=2.6pt of Bl5] (Bl6) {Procedural generation and 3D reconstruction \taxcite{Chen2024SpatialVLMEV,Yuan2024RoboPointAV}};
\node[leaf, fill=taxbluL, below=2.6pt of Bl6] (Bl7) {Model-generated \taxcite{Batra2025SpatialThinkerR3,Shen2025FineGrainedPO}};
\node[leaf, fill=taxbluL, below=2.6pt of Bl7] (Bl8) {Human supervised \taxcite{Chen2025Robo2VLMVQ,Huang2025SpatialDISEAU}};
\node[cat, fill=taxbluM, below=7pt of Bl8] (Bc3) {Data Modalities};
\node[leaf, fill=taxbluL, below=2.6pt of Bc3] (Bl9) {Vision and language \taxcite{Song2025RoboSpatialTS,Chen2025Robo2VLMVQ}};
\node[leaf, fill=taxbluL, below=2.6pt of Bl9] (Bl10) {Depth and 3D information \taxcite{Liu2025SSRED,Zhang2025Open3DVQAAB}};
\node[leaf, fill=taxbluL, below=2.6pt of Bl10] (Bl11) {Bounding boxes and segmentation masks \taxcite{Cheng2024SpatialRGPTGS,Ogezi2025SpaREES}};
\node[leaf, fill=taxbluL, below=2.6pt of Bl11] (Bl12) {Video, trajectories, and keypoints \taxcite{Wang2025VideoSTRRM,Yang2025VisionThinkSA}};
\node[cat, fill=taxbluM, below=7pt of Bl12] (Bc4) {Source Environment};
\node[leaf, fill=taxbluL, below=2.6pt of Bc4] (Bl13) {Robotics and autonomous driving \taxcite{Song2025RoboSpatialTS,Yuan2024RoboPointAV}};
\node[leaf, fill=taxbluL, below=2.6pt of Bl13] (Bl14) {General purpose \taxcite{Deng2025InternSpatialAC,Cheng2024SpatialRGPTGS}};
\node[leaf, fill=taxbluL, below=2.6pt of Bl14] (Bl15) {Simulated indoor environments \taxcite{Yuan2024RoboPointAV}};
\node[leaf, fill=taxbluL, below=2.6pt of Bl15] (Bl16) {Aerial and drone-view video \taxcite{Sun2025SpaceVistaAV,Zhang2025Open3DVQAAB}};

\draw[spine] ([xshift=3pt]hB.south west) |- (Bc1.west);
\draw[spine] ([xshift=3pt]hB.south west) |- (Bc2.west);
\draw[spine] ([xshift=3pt]hB.south west) |- (Bc3.west);
\draw[spine] ([xshift=3pt]hB.south west) |- (Bc4.west);
\node[cat, fill=taxperM, anchor=north west] (Cc1) at ([xshift=8pt,yshift=-7pt]hC.south west) {What Is Measured};
\node[leaf, fill=taxperL, below=2.6pt of Cc1] (Cl1) {Target-skill performance decomposed by task type \taxcite{Zhang2025FromFT,Huang2025SpatialDISEAU}};
\node[leaf, fill=taxperL, below=2.6pt of Cl1] (Cl2) {General-capability retention on vision--language benchmarks \taxcite{Ogezi2025SpaREES,Li2025SpatialLadderPT}};
\node[cat, fill=taxperM, below=7pt of Cl2] (Cc2) {Metrics by Answer Type};
\node[leaf, fill=taxperL, below=2.6pt of Cc2] (Cl3) {Accuracy for categorical answers (e.g., left/right, yes/no) \taxcite{Zhang2025FromFT,Huang2025SpatialDISEAU}};
\node[leaf, fill=taxperL, below=2.6pt of Cl3] (Cl4) {Tolerance-based scoring for metric answers (e.g., distance, size) \taxcite{Cheng2024SpatialRGPTGS,Ouyang2025SpaceRRM}};
\node[leaf, fill=taxperL, below=2.6pt of Cl4] (Cl5) {LLM-as-judge for open-ended answers \taxcite{Liu2025SSRED,Deng2025InternSpatialAC}};
\node[leaf, fill=taxperL, below=2.6pt of Cl5] (Cl6) {Geometric and end-to-end task success for action answers \taxcite{Song2025RoboSpatialTS,Yuan2024RoboPointAV}};
\node[leaf, fill=taxperL, below=2.6pt of Cl6] (Cl7) {Human baselines for calibration \taxcite{Chen2025Robo2VLMVQ,Huang2025SpatialDISEAU}};
\node[cat, fill=taxperM, below=7pt of Cl7] (Cc3) {Generalization Testing};
\node[leaf, fill=taxperL, below=2.6pt of Cc3] (Cl8) {In-domain vs.\ out-of-domain benchmarks \taxcite{Li2025SpatialLadderPT,Wu2025ReinforcingSR}};
\node[leaf, fill=taxperL, below=2.6pt of Cl8] (Cl9) {Robustness probes \taxcite{Cheng2024SpatialRGPTGS,Song2025RoboSpatialTS,Ouyang2025SpaceRRM}};
\node[cat, fill=taxperM, below=7pt of Cl9] (Cc4) {Open Problems};
\node[leaf, fill=taxperL, below=2.6pt of Cc4] (Cl10) {LLM judges unvalidated on spatial content \taxcite{Liu2025SSRED,Cheng2024SpatialRGPTGS}};
\node[leaf, fill=taxperL, below=2.6pt of Cl10] (Cl11) {Conventional, unjustified tolerance thresholds (e.g., $\pm$25\%) \taxcite{Cheng2024SpatialRGPTGS}};
\node[leaf, fill=taxperL, below=2.6pt of Cl11] (Cl12) {Shared train--eval construction pipelines \taxcite{Liu2025SSRED,Deng2025InternSpatialAC}};
\draw[spine] ([xshift=3pt]hC.south west) |- (Cc1.west);
\draw[spine] ([xshift=3pt]hC.south west) |- (Cc2.west);
\draw[spine] ([xshift=3pt]hC.south west) |- (Cc3.west);
\draw[spine] ([xshift=3pt]hC.south west) |- (Cc4.west);
\end{tikzpicture}%

    \end{adjustbox}
    \caption{Taxonomy of data-centric spatial reasoning in MLLMs, organized by dataset design, construction, and evaluation.}
    \label{fig:taxonomy}
\end{figure}
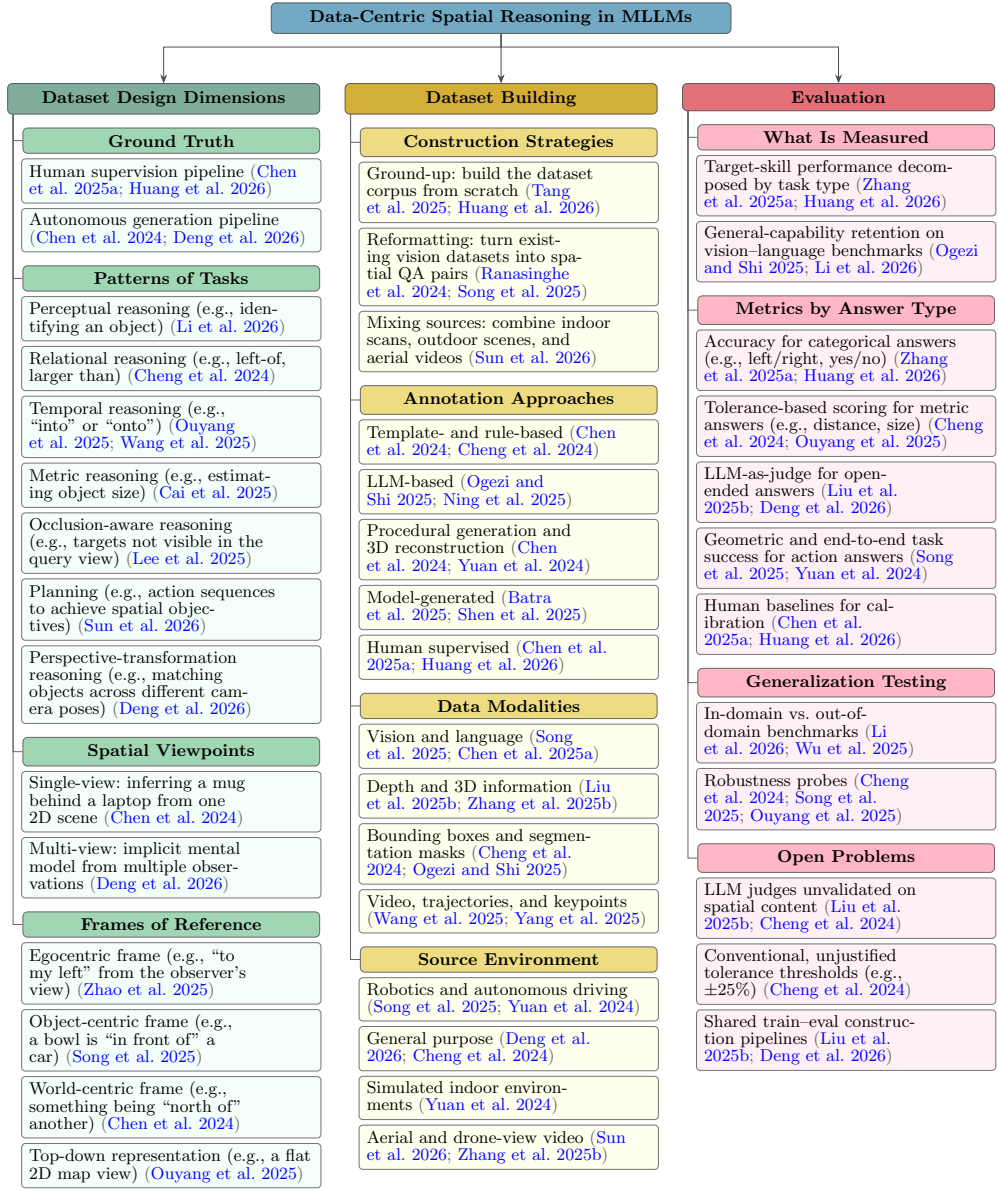

\begin{itemize}

\item A new taxonomy of selected datasets for MLLM fine-tuning for spatial reasoning (Figure~\ref{fig:taxonomy}).
\item A critical analysis of the dataset construction pipelines, and how synthetic and human annotations are blended to produce training datasets.
\item An audit of evaluation protocols across the surveyed datasets, revealing inconsistent scoring criteria, tolerance thresholds, and benchmark construction that limit cross-work comparability.
\item A proposal outlining insights, gaps, and future directions, including unexplored spatial tasks, scalability issues, and cross-modal and cross-domain generalization.

\end{itemize}

The rest of this survey follows the three branches of the taxonomy in Figure~\ref{fig:taxonomy}. Section 2 provides background context and reviews related surveys. Section 3 examines dimensions of dataset design, including ground truth, task patterns, viewpoints, and reference frames. Section 4 covers dimensions of dataset creation, including construction methods, annotation approaches, modality, and source environments. Section 5 addresses aspects of evaluation, including what is evaluated, how answers are scored, and how generalization is evaluated. Section 6 makes comparisons across datasets and identifies patterns, open problems, and opportunities for future work.

\section{Background}
\label{sec:background}

\subsection{Spatial Awareness and Visual Reasoning}
Spatial awareness refers to the perceptual and cognitive abilities to derive spatial information from visual input and to maintain an internal representation of the spatial configuration of objects in a scene (such as their location or relative positioning)~\cite{Ranasinghe2024LearningTL}. The presence of an internal representation that is independent of the observer has been proposed across many studies in cognitive science~\cite{OKeefe1979TheHA}.  Internal representations have been linked to spatial manipulation and mental simulation tasks, which we refer to as spatial cognition~\cite{Hegarty2004MechanicalRB}. While these two terms are often used interchangeably in the literature, we use the distinction here to help organize the survey of these capabilities, as some works use spatial cognition to refer to what we classify as spatial awareness~\cite{Cheng2024SpatialRGPTGS,Zhao2025EmbodiedRCF}.
This forms a foundational capability for visual reasoning and interaction with the environment~\cite{Piaget2013TheCC}. 

Visual reasoning is commonly assumed to build on spatial awareness to perform high-level reasoning over visual input. It uses perceptual and spatially aware knowledge to draw inferences, make comparisons, and answer questions that go beyond observing visual scenes. Prior work has shown a deficiency in language-only reasoning for spatial reasoning~\cite{Li2024AdvancingSR}. However, merely reasoning over visual information is not sufficient to overcome these shortcomings. There are still many failures in spatial relations in vision-language models~\cite{Kamath2023WhatsUW}. These shortcomings emphasize the need for specialized post-training and fine-tuning, which is the focus of this survey, in order to correctly align MLLMs with spatial concepts.

\subsection{Related Surveys} \label{relatedSurvey}
Recent surveys have investigated spatial reasoning in MLLMs from complementary angles. From a general multimodal perspective, \citet{Bai2024ASO} offer a dataset-level, modality-agnostic survey of datasets and benchmarks for evaluating MLLMs, though not specifically with respect to spatial reasoning. From the narrower focus of spatial reasoning, \citet{Liu2025SpatialRI} survey spatial reasoning tasks, benchmarks, and methods from a cognitive science view: they propose a taxonomy based on cognitive complexity, and map existing benchmarks to it, as well as model architectures and algorithms, and survey how they are (or can be) evaluated and improved. \citet{Zheng2025MultimodalSR} focus on model architectures, capabilities, and post-training extending the discussion from 2D VLMs to 3D and embodied AI environments, and propose benchmarks for evaluating spatial reasoning. \citet{Feng2025ASO} broaden the scope along a different axis, surveying LLM-powered spatial intelligence across embodied, urban, and earth scales through the lens of human spatial cognition. \citet{Ma2024WhenLS} focus on 3D settings and survey approaches to connect LLMs to 3D information, as well as provide a meta-analysis on the current state of 3D-LLMs. Similarly, \citet{Zha2025HowTE} survey methods for enabling LLMs to understand 3D space, categorizing these approaches into image-based, point cloud-based, and hybrid methods. In addition to these method-focused surveys, \citet{Yu2025HowFA} survey visual spatial understanding in VLMs via benchmarks. They consolidate nearly 20 open-source datasets into a benchmark suite (SIBench) that highlights a large disconnect between visual perception and spatial understanding. Lastly, \citet{Liu2025SpatialII} provide a comprehensive survey of spatial understanding in VLMs. They cover definitions of spatial tasks, error analysis, methods for improvement, evaluation techniques, and use cases. Furthermore, they propose a taxonomy based on intervention methods and support this with an extensive empirical study. To complement these works, our survey focuses specifically on the construction and comparison of selected spatial post-training datasets, their annotation schemes, and how model performance is evaluated.


\section{Spatial Reasoning Dataset Design Dimensions}
\label{sec:spatial_reasoning_tasks}

Datasets used for fine-tuning foundation models to improve spatial understanding typically include paired visual inputs with textual annotations encoding different spatial attributes such as object relations, depth and distances, geometric structures, and layouts~\citet{Liu2025SpatialII}. When the annotations are in the form of question-answer pairs, they can be used to create a visual question answering (VQA) dataset that is commonly used for fine-tuning. In supervised instruction tuning, these models are fine-tuned as conditional text generators given visual and textual inputs for next-token prediction with multimodal supervision, so VQA is a model-agnostic format for fine-tuning datasets \citep{Liu2023VisualIT,Li2023BLIP2BL}. However, some datasets extend beyond conventional VQA by feeding different modalities to the model or encoding their cues into textual annotations (see Section~\ref{sec:datasets} for more details). 

\subsection{Ground Truth Spatial Reasoning}
While VLMs are typically pre-trained on large-scale, unstructured multimodal corpora through self-supervised objectives, their post-training effectiveness depends heavily on the quality of ground-truth annotated datasets.
Ground truth in machine learning, in general, is an external reference signal for supervising the learning process~\cite{Jaton2021AssessingBR}, and in LLM reasoning, it is a set of external signals used to evaluate the reasoning trace~\cite{Lee2025EvaluatingSR}. In spatial reasoning, this definition can be extended to a set of structured reference specifications of spatial information, such as object relations and locations, depth, and compositional plans, used to train or evaluate a model’s performance given a visual or multimodal input~\cite{Johnson2017CLEVRAD}. From a data-centric perspective, building ground-truth spatial understanding is challenging since most spatial attributes are implicit, and their availability and reliability depend on the source of information. Therefore, it has to be created through a carefully designed dataset generation pipeline, either autonomously or under human supervision.

\subsection{Patterns of Tasks: Perception to Reasoning}
To train VLMs for spatial understanding, datasets derive a spatial learning signal from a ground truth source. This is typically done in practice by defining task types that specify what the spatial reasoning is meant to teach. To provide a view of some recurring patterns, at the simplest level, many datasets define a task that relates to perception, such that the model’s reasoning is expected to relate that spatial language to visual evidence, such as identification, localization, or presence of an object. SpatialLadder~\cite{Li2025SpatialLadderPT} defines this approach in the context of a hierarchy of tasks, with perceptual reasoning defined as the first level of a hierarchical curriculum, claiming that higher-level forms of reasoning can only be reliably performed if perception is accurate.
Relational reasoning tasks require the model to reason about the relationships between objects in the image, rather than relying on perception alone. This typically takes the form of pairs of objects in the scene, which are annotated with qualitative relationships (e.g., left-of, above, behind) or comparative relationships (e.g., closer to, larger than). In this case, ground truth is typically derived from 3D scene metadata or depth of objects in the scene~\cite{Cheng2024SpatialRGPTGS}. A variation on this task type is perspective-transformation reasoning, which requires reasoning in relation to different perspectives, such as the camera’s motion or matching objects in images with different camera poses~\cite{Deng2025InternSpatialAC}.
Temporal and dynamic reasoning builds upon this with sequences: one can use static relations to reason about something being \enquote{above} another, but dynamic relations such as \enquote{into} or \enquote{onto} require motion in their creation. As a result, ground truth for dynamic reasoning can be obtained from video sources~\cite{Ouyang2025SpaceRRM,Wang2025VideoSTRRM}. However, dynamic reasoning is not exclusively obtained from video: Spatial-DISE also uses dynamic reasoning on objects from static images~\citep{Huang2025SpatialDISEAU}. Thus, motion of objects, motion of the camera, and a hypothetical transformation should be clearly specified for dynamic reasoning tasks.

Metric reasoning requires models to estimate quantities, such as absolute distances or object sizes. Ground truth is obtained either from depth sensors~\cite{Cai2025SpatialBotPS} or from depth maps estimated during data generation ~\cite{Cheng2024SpatialRGPTGS, Chen2024SpatialVLMEV}.
Less common tasks include occlusion-aware reasoning from visibility masks~\cite{Lee2025SpatialMosaicAM} and planning tasks that indicate action sequences to achieve spatial objectives~\cite{Sun2025SpaceVistaAV}.
These task categories typically fall into one of the splits defined in Section~\ref{sec:background}. For perceptual and metric tasks, the model must use the input to gain knowledge about spatial relationships, or possess spatial awareness. For perspective-transformation, dynamic, occlusion-aware and planning tasks, the model must manipulate or extrapolate from an already-existing representation, or possess spatial cognition. Relational reasoning straddles this divide as reading a relationship from an observed pair can be considered perceptual, but comparing across objects in a transformation cannot. Most datasets include examples from both sides of this divide.

\subsection{Spatial Viewpoints and Frames of Reference}
Beyond task selection, spatial reasoning datasets also need to define a viewpoint for observation. Many use a single-view approach, with datasets like SpatialVLM~\cite{Chen2024SpatialVLMEV}, and SpatialRGPT~\cite{Cheng2024SpatialRGPTGS} testing whether a model can infer a 3D relation, for example, a mug being behind a laptop, from a singular observation of a 2D scene. On the other hand, InternSpatial~\cite{Deng2025InternSpatialAC} builds off this notion of single-view and multi-view, providing spatial supervision such that models need to connect information from multiple observations of the same spatial scene rather than relying on just one view.

The way spatial information is observed influences how it is represented and also the choice of reference frame influences what spatial relations can even be expressed. In general, spatial relations are defined relative to a reference frame. In an egocentric reference frame, relations between objects are expressed relative to the observer. This is especially relevant to embodied agents~\cite{Zhao2025EmbodiedRCF}. In a world-centric reference frame, relations between objects are expressed in world coordinates external to the observer. For example, an object can be expressed as being above another object irrespective of the location of the observer in the world~\cite{Chen2024SpatialVLMEV}. In an object-centric reference frame, spatial information is expressed relative to the object itself. This distinction of frames has an influence on correct answers; for example, in RoboSpatial~\cite{Song2025RoboSpatialTS}, a bowl can be in front of a car from a third-person perspective but behind the car in an egocentric perspective. Thus, for the same scene, different answers can be correct depending on the reference frame. Finally, some datasets employ a flat 2D map view of the environment, as in SpaceR-151k~\cite{Ouyang2025SpaceRRM}.

Spatial reasoning datasets that ask the models to reason about spatial relations in multiple frames of reference, such as SenseNova-SI-8M~\cite{Cai2025ScalingSI}, help the models to develop an understanding of how the same spatial relation can be described differently depending on the chosen frame of reference.

Table~\ref{tab:spatial_reasoning} records the viewpoints, reference frames, and spatial representation of each selected dataset: single-view settings dominate compared to multi-view settings. Egocentric framing is the most common frame, and four datasets adopt a top-down representation.


%
%


\newcommand{\na}{--}

\newcommand{\mixedsize}[1]{\mbox{#1\ensuremath{^{\Sigma}}}}




\newcommand{\singleview}{single}
\newcommand{\multiview}{multi}
\newcommand{\topdownmap}{topdown}
\newcommand{\objectcentric}{object}
\newcommand{\egocentric}{ego}
\newcommand{\worldcentric}{world}



\newcommand{\video}{vid}
\newcommand{\rgb}{rgb}
\newcommand{\boundingbox}{bbox}
\newcommand{\keypoint}{kpt}
\newcommand{\segmentation}{seg}

\newcommand{\threedimensional}{3d}
\newcommand{\depthmap}{dmap}
\newcommand{\pointcloud}{ptcl}
\newcommand{\threedimensionalmesh}{3dmesh}
\newcommand{\camerapose}{cpose}

\newcommand{\naturallanguage}{nl}
\newcommand{\trajectory}{traj}



\newcommand{\allm}{llm}
\newcommand{\aprocedural}{proc}
\newcommand{\atemplaterule}{tmpl}
\newcommand{\aother}{other}

\newcommand{\hsemisupervision}{semi}
\newcommand{\hfullsupervision}{full}

\newcommand{\mcomputervision}{cv}

\newcommand{\mreinforcementlearning}{rl}


\begin{sidewaystable}[p]
    \centering
    \caption{
    Spatial reasoning post-training datasets and dataset collections.
    Several entries also serve as benchmarks; these are included where a training split accompanies the benchmark.
    Codes are defined in the legend below.
    }
    \label{tab:spatial_reasoning}
    \fontsize{7}{8}\selectfont
    \setlength{\tabcolsep}{2pt}
    \renewcommand{\arraystretch}{0.85}

        \begin{tabular}{
            p{3.25cm}  
            p{2.88cm}  
            p{1.69cm}  
            p{3.38cm}  
            p{4.51cm}  
            p{2.88cm}  
            }
        \toprule
        \textbf{Citation}
        & \textbf{Name}
        & \textbf{Scale}
        & \textbf{View}
        & \textbf{Modality}
        & \textbf{Annotation}\\
        \midrule

        \cite{Song2025RoboSpatialTS}
        & RoboSpatial
        & \mixedsize{4M}
        & \singleview, \objectcentric, \egocentric, \worldcentric
        & \rgb, \boundingbox, \depthmap, \pointcloud, \camerapose, \naturallanguage
        & \allm, \aprocedural, \atemplaterule
        \\

        \cite{Chen2025Robo2VLMVQ}
        & Robo2VLM-1
        & 685k
        & \multiview, \egocentric, \worldcentric
        & \rgb, \segmentation, \naturallanguage, \trajectory
        & \atemplaterule, \hsemisupervision
        \\

        \cite{Ning2025EnhancingSR}
        & 3D ReasonSeg
        & 29.2k
        & \worldcentric
        & \segmentation, \pointcloud, \naturallanguage
        & \allm, \atemplaterule, \hsemisupervision
        \\

        \cite{Liu2025SSRED}
        & SSR-COT
        & 1.2M
        & \singleview, \worldcentric
        & \rgb, \depthmap, \naturallanguage
        & \allm, \atemplaterule, \aother
        \\

        \cite{Deng2025InternSpatialAC}
        & InternSpatial
        & 12M
        & \singleview, \multiview, \egocentric
        & \rgb, \boundingbox, \segmentation, \naturallanguage
        & \allm, \aprocedural, \atemplaterule, \hsemisupervision
        \\

        \cite{Ranasinghe2024LearningTL}
        & Localize\ensuremath{^\dagger}
        & 200k+
        & \singleview
        & \video, \rgb, \boundingbox, \naturallanguage
        & \allm, \atemplaterule, \hsemisupervision
        \\

        \cite{Zhang2025Open3DVQAAB}
        & Open3D-VQA
        & \mixedsize{74.2k}
        & \singleview, \topdownmap, \objectcentric, \egocentric, \worldcentric
        & \rgb, \boundingbox, \segmentation, \depthmap, \pointcloud, \camerapose, \naturallanguage
        & \allm, \aprocedural, \atemplaterule, \hsemisupervision
        \\

        \cite{Yuan2024RoboPointAV}
        & RoboPoint-1432K
        & 1432k
        & \singleview, \egocentric, \worldcentric
        & \rgb, \boundingbox, \keypoint, \depthmap, \naturallanguage
        & \aprocedural, \atemplaterule
        \\

        \cite{Ogezi2025SpaREES}
        & SpaRE
        & \mixedsize{3.8M}
        & \singleview, \objectcentric
        & \rgb, \naturallanguage
        & \allm, \aprocedural, \hsemisupervision
        \\

        \cite{Tang2025SparkleMB}
        & Sparkle
        & \mixedsize{180k}
        & \singleview, \topdownmap, \worldcentric
        & \rgb, \naturallanguage
        & \aprocedural, \atemplaterule
        \\

        \cite{Huang2025SpatialDISEAU}
        & Spatial-DISE-12K
        & 12.4k
        & \singleview, \multiview, \objectcentric
        & \rgb, \naturallanguage
        & \aprocedural, \atemplaterule, \hsemisupervision
        \\

        \cite{Ouyang2025SpaceRRM}
        & SpaceR-151k
        & 151k
        & \singleview, \topdownmap, \egocentric
        & \video, \rgb, \boundingbox, \pointcloud, \naturallanguage
        & \aprocedural, \atemplaterule
        \\

        \cite{Batra2025SpatialThinkerR3}
        & STVQA-7K
        & 7.6k
        & \singleview, \egocentric
        & \rgb, \boundingbox, \naturallanguage
        & \allm, \hfullsupervision
        \\

        \cite{Shen2025FineGrainedPO}
        & Open Spatial Reasoning
        & \mixedsize{900k}
        & \singleview, \objectcentric
        & \rgb, \segmentation, \depthmap, \naturallanguage
        & \allm, \aprocedural
        \\

        \cite{Wu2025ReinforcingSR}
        & VILASR\ensuremath{^\dagger}
        & \mixedsize{80.8k}
        & \singleview, \multiview
        & \video, \rgb, \boundingbox, \naturallanguage
        & \allm, \aprocedural, \atemplaterule
        \\

        \cite{Sarch2025GroundedRL}
        & ViGoRL
        & \mixedsize{86.6k}
        & \singleview, \objectcentric
        & \rgb, \naturallanguage
        & \allm, \aother
        \\

        \cite{Chen2025PerceptionBR}
        & \na
        & \mixedsize{409k}
        & \singleview, \egocentric 
        & \rgb, \naturallanguage
        & \allm, \atemplaterule, \hsemisupervision, \hfullsupervision
        \\

        \cite{Zhao2025EmbodiedRCF}
        & \na
        & 5.4k
        & \egocentric, \worldcentric
        & \video, \rgb, \keypoint, \naturallanguage, \trajectory
        & \na
        \\

        \cite{Chen2025ThinkW3}
        & \na
        & \mixedsize{247k}
        & \multiview, \egocentric
        & \rgb, \naturallanguage
        & \allm
        \\

        \cite{Cai2025SpatialBotPS}
        & SpatialQA\ensuremath{^\dagger}
        & 746k
        & \singleview, \objectcentric, \egocentric
        & \rgb, \boundingbox, \depthmap, \naturallanguage, \trajectory
        & \allm, \aprocedural, \atemplaterule, \hsemisupervision
        \\

        \cite{Li2025SpatialLadderPT}
        & SpatialLadder-26K
        & 26.6k
        & \singleview, \multiview, \egocentric, \worldcentric
        & \video, \rgb, \boundingbox, \naturallanguage
        & \aprocedural, \atemplaterule
        \\

        \cite{Cheng2024SpatialRGPTGS}
        & OSD
        & 8.7M
        & \singleview, \egocentric, \worldcentric 
        & \rgb, \boundingbox, \segmentation, \depthmap, \naturallanguage
        & \allm, \aprocedural, \atemplaterule
        \\

        \cite{Chen2024SpatialVLMEV}
        & \na
        & 2B
        & \singleview, \egocentric, \worldcentric
        & \rgb, \boundingbox, \segmentation, \depthmap, \pointcloud, \naturallanguage
        & \aprocedural, \atemplaterule
        \\

        \cite{Wang2025VideoSTRRM}
        & STV-205k
        & 205k
        & \multiview
        & \video, \rgb, \boundingbox, \naturallanguage
        & \aprocedural, \atemplaterule, \hfullsupervision
        \\

        \cite{Sun2025SpaceVistaAV}
        & SpaceVista-1M
        & 1M
        & \multiview, \egocentric, \worldcentric
        & \video, \rgb, \boundingbox, \keypoint, \segmentation, \naturallanguage
        & \allm, \aprocedural, \atemplaterule, \hsemisupervision
        \\

        \cite{Zhang2025FromFT}
        & SPAR-7M
        & 7M
        & \singleview, \multiview, \egocentric
        & \video, \rgb, \boundingbox, \keypoint
        & \allm, \aprocedural, \atemplaterule
        \\

        \cite{Lee2025SpatialMosaicAM}
        & SpatialMosaic
        & 2M
        & \multiview, \egocentric
        & \rgb, \boundingbox, \depthmap, \pointcloud, \camerapose, \naturallanguage
        & \aprocedural, \atemplaterule
        \\

        \cite{Cai2025ScalingSI}
        & SenseNova-SI-8M
        & 8.5M
        & \singleview, \multiview, \topdownmap, \objectcentric, \egocentric, \worldcentric
        & \video, \rgb, \boundingbox, \pointcloud, \naturallanguage
        & \aprocedural, \aother
        \\

        \bottomrule
    \end{tabular}

    \vspace{4pt}
    \begin{minipage}{\linewidth}
        \footnotesize
        \setlength{\parindent}{0pt}

        \textbf{Name.} \ensuremath{^\dagger} marks dataset families reported under a single name: Localize (Localize-Instruct-200K, Localize-ActivityNet); VILASR (VILASR-ColdStart-33k, VILASR-RRS-8k, VILASR-RL-40k); SpatialQA (SpatialQA, SpatialQA-E). A dash (\na) denotes not reported, here and in every column.

        \textbf{Scale.} Rounded QA/example counts unless marked;
        \textit{B}/\textit{M}/\textit{k}: billions/millions/thousands;
        \textit{+}: lower bound. Superscript \ensuremath{\Sigma}: survey-computed
        sum across visual/annotation records or stage collections.

        \textbf{View.} Combines viewpoint with frame of reference. Viewpoint: single-view (\textit{\singleview}), multi-view (\textit{\multiview}). Frame: object-centric (\textit{\objectcentric}), egocentric (\textit{\egocentric}), world-centric (\textit{\worldcentric}). Top-down (\textit{\topdownmap}) denotes a flat 2D map representation.

        \textbf{Modality.} Vision: video (\textit{\video}), RGB (\textit{\rgb}), bounding box (\textit{\boundingbox}), keypoint (\textit{\keypoint}), segmentation (\textit{\segmentation}). Geometry: depth map (\textit{\depthmap}), point cloud (\textit{\pointcloud}), camera pose (\textit{\camerapose}). Text: natural language (\textit{\naturallanguage}), trajectory (\textit{\trajectory}).

        \textbf{Annotation.} How the fixed training corpus was annotated. Automated: LLM-based (\textit{\allm}), procedural or 3D reconstruction (\textit{\aprocedural}), template- or rule-based (\textit{\atemplaterule}), supervision derived from auxiliary vision models or from the model's own outputs (\textit{\aother}). Human involvement: partial (\textit{\hsemisupervision}), full (\textit{\hfullsupervision}).
    \end{minipage}
\end{sidewaystable}

\section{Dataset Building}
\label{sec:datasets}

\begin{sidewaysfigure}
    \centering
    \includegraphics[width=\linewidth]{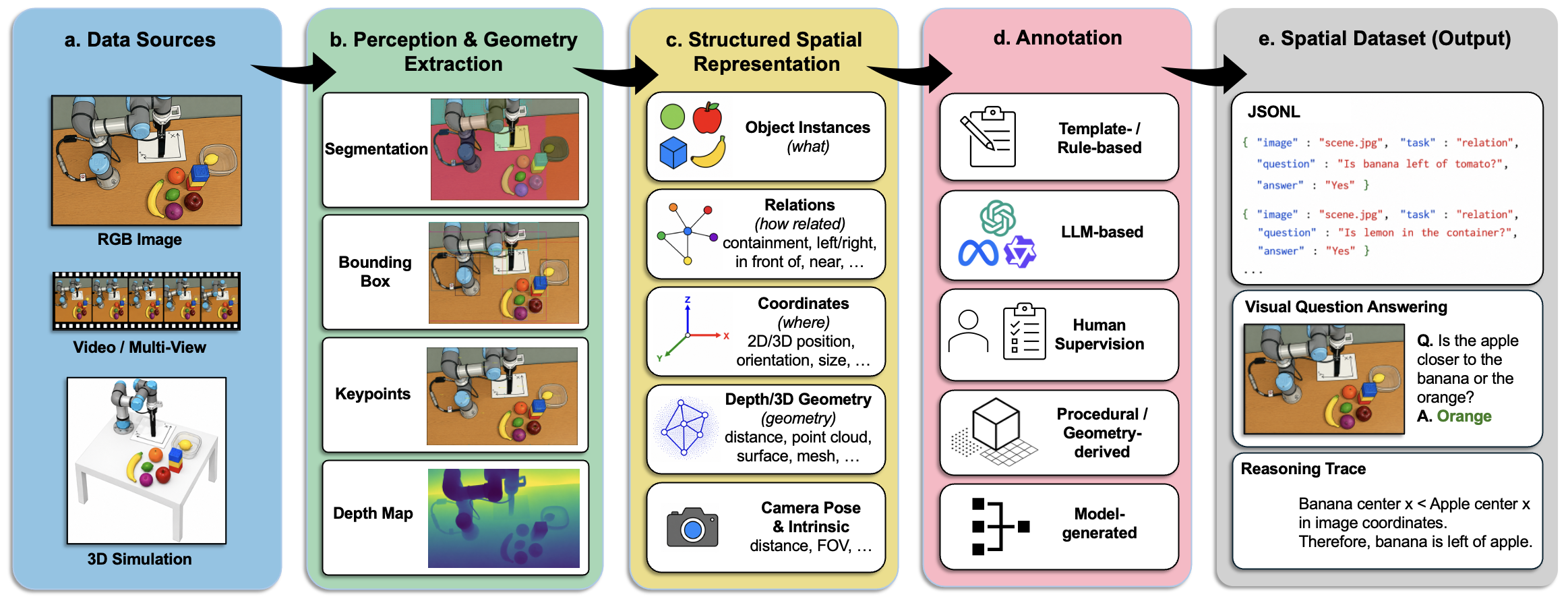}
    \caption{Overview of the common pipeline across the literature for building spatial reasoning datasets. Where needed, raw visual data are first processed by perception and geometry models to extract spatial cues. Then they are fed into structured spatial representations. These representations provide the supervision signal used by different annotation approaches to generate the final spatial training data, which includes VQA pairs and optional reasoning traces.}
    \label{fig:pipeline}
\end{sidewaysfigure}

This section examines how spatial reasoning datasets for VLMs are built in practice. We review how datasets are constructed or adapted from existing data resources, how spatial information is annotated, which modalities are supported, and which environments serve as the sources of data.
Most of these datasets use a similar data pipeline as shown in Figure~\ref{fig:pipeline}. Raw data may include single RGB images, video or multi-view images, or 3D simulations. Using perception and geometry models these inputs are transformed into depth, keypoints, bounding boxes and segmentation masks. In other pipelines, geometry can be derived from captured 3D data or the states of a simulator. Spatial knowledge can also come from human annotations or captions. Then the masks, visual labels and maps that are extracted, are passed into a more structured spatial representation. At this step, the information undergoes additional transformations to add semantic, spatial and 3D information by having organizing geometry, camera parameters, coordinates, relationships, object instances and spatial information. Now, the spatial supervision signal is ready to be used by annotation methods. The annotation stage then turns this signal into training data through templates, LLM-generated rationales and captions, human supervision, procedural computation, or by the model-generated labels. Annotation produces VQA pairs and, in some cases, reasoning traces. The following subsections examine each of these stages with more details. 


\subsection{Dataset Construction Strategies}
Most of the surveyed spatial reasoning datasets do not build their visual data collection from the ground up, but instead extend general-purpose vision datasets, reformatting their annotations into spatial QA style, and allowing for spatial supervision to be extended to existing datasets and scaled without new data collection. For instance, \citet{Ranasinghe2024LearningTL} convert bounding boxes to instruction-based QA with coordinate answers and scale with automated data generation. Other work extends even further, building on existing 2D or 3D datasets and extending them with automatically computed spatial annotations (e.g., robot-relevant signals such as free space), and then generating a massive scale of spatial QA pairs from those~\cite{Song2025RoboSpatialTS}. Some approaches leverage multiple datasets to maximize coverage and diversity. For instance, SpaceVista-1M mixes sources in their curated dataset, combining indoor scans, outdoor scenes, and aerial videos~\cite{Sun2025SpaceVistaAV}. Similarly, SPAR-7M~\cite{Zhang2025FromFT} reformats existing richly annotated 3D indoor scene datasets into 7M QA pairs spanning 33 task types across single-view, multi-view, and video inputs, using scene geometry to generate supervision that ranges from 2D perception to 3D-grounded reasoning. SenseNova-SI~\cite{Cai2025ScalingSI} takes this aggregation strategy further by combining 4M samples from different open-source datasets under a unified spatial-capability taxonomy, then generating another 4.5M QA pairs from richly annotated 3D sources (ScanNet, ScanNet++, SUN RGB-D, Matterport3D, and others) to cover less represented capabilities such as perspective-taking, for a total of 8.5M pairs.
Few of the recent datasets are built from scratch, which are typically designed for a specific target task or for benchmarking rather than training at scale. Sparkle~\citep{Tang2025SparkleMB}, for instance, generates a small-scale synthetic dataset that teaches basic spatial reasoning capabilities (such as direction, distance, and localization) with the goal of generalizing to harder tasks. Spatial-DISE~\cite{Huang2025SpatialDISEAU} has a similar focus on simple abilities, but is mainly designed as a hybrid benchmark dataset, combining real-world problems with synthetic scenes.

\subsection{Annotation Approaches}
The annotation approach used to create a fine-tuning dataset inherently constrains what spatial concepts a VLM dataset can directly supervise. The annotation of the training data for spatial understanding, whether as coordinate pairs, natural language text, reasoning paths, or the outputs of a model, exposes models to the \enquote{ground truth} spatial structure that may be underrepresented in pre-training. Additionally, annotation approaches vary in their quality, but all come to a quality-quantity equilibrium with varying scale of supervision, precision, coverage, and human involvement. We examine annotation approaches that lead to the creation of fine-tuning datasets, with a spectrum from automated to human. Across most annotation approaches, the pre-trained perception model extracts object- and geometry-level structure from the visual input using auxiliary models (e.g., object detectors, depth or geometry estimators), which downstream templates, LLMs, procedural rules, or learning objectives then leverage to generate spatial supervisions~\citep{Chen2024SpatialVLMEV,Cheng2024SpatialRGPTGS,Song2025RoboSpatialTS}.
The Annotation column of Table~\ref{tab:spatial_reasoning} lists the annotation method for each dataset. These annotation methods represent potential elements that can appear within the same pipeline and are not necessarily mutually exclusive. Instead, they can be combined at various stages of annotation.

\subsubsection{Template-Based and Rule-Based Annotation}
\begin{figure}[htb]
    \centering
    \includegraphics[width=0.99\linewidth]{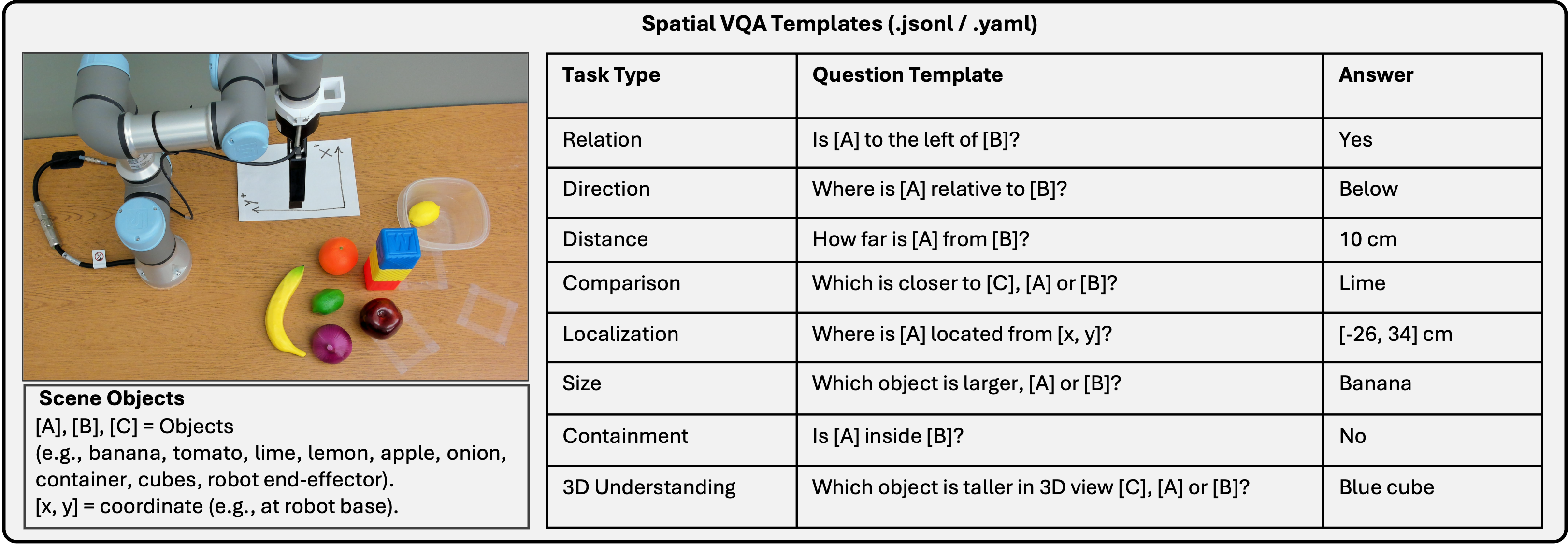}
    \caption{Template- and rule-based spatial VQA annotation (example from a robot-manipulation task). Explicit spatial rules define ground-truth and reference relations over structured scene representations, which are verbalized through controlled templates to produce scalable supervision. These are also conditional on precise object bindings and coordinate conventions.}
    \label{fig:vqa_templates}
\end{figure}

In template-based annotation, a rule and a sentence are paired together. The rule is applied to a structured scene and produces an answer to a relation (e.g., the location of one object relative to another or the distance between two objects). The template then expresses that relation as a question. Because the mapping goes from relation to text, the answer is generated prior to the question and a single rule can be applied across an entire corpus (see Figure~\ref{fig:vqa_templates}). SpatialVLM~\citep{Chen2024SpatialVLMEV} defines 38 types of spatial questions that can express either qualitative or quantitative relations. Each question and answer type is expressed using a small number of templates to create a large and balanced corpus of questions and answers. Similarly, SpatialRGPT~\citep{Cheng2024SpatialRGPTGS} converts 3D scene graphs into region-level questions and answers. Hand-crafted templates are used to ask questions regarding object dimensions (height/width) and qualitative relations (left/right or direction), as well as quantitative metrics (distance). While this system uses a mix of approaches, these templates are the foundation of its large collection of spatial questions and answers. InternSpatial~\citep{Deng2025InternSpatialAC} develops an automated system that converts varied visual inputs into spatial questions and answers. Each question is then converted into visual and textual references using template-based methods that utilize 3D view space. SpatialLadder~\citep{Li2025SpatialLadderPT} enforces consistency across questions and answers by using a small number of question templates across modalities and tasks. Geometric consistency is maintained by leveraging 3D reconstructions for template instantiations.
Sparkle~\citep{Tang2025SparkleMB} decomposes 2D spatial reasoning into three components: direction, distance, and localization, and designs prompt templates for each, showing that models trained on these basic templates generalize to other composite tasks (e.g., shortest path problems).

One drawback of template-based approaches is linguistic inflexibility. InternSpatial~\citep{Deng2025InternSpatialAC} identifies limitations and addresses them by introducing LLMs to paraphrase and expand template-generated questions, before the QA pairs are instantiated, without losing their spatial semantics. However, this issue is not equally problematic across all scenarios. When using GPT-4 to rephrase the evaluation questions and answers in SpatialRGPT-Bench, SpatialRGPT-7B’s qualitative score (concerning rephrasing procedure) only drops by 1.8 points (from 91.8 to 90.0)~\citep{Cheng2024SpatialRGPTGS}. A more general issue with templates is that they can only provide supervision data for spatial relationships that are present in the source representation. This design choice marks a natural transition to annotation pipelines in which LLMs play a generative role.

\subsubsection{LLM-Based Annotation}

LLMs offer a flexible means of introducing linguistic variation into systems to create questions that are difficult to define with hand-written templates. SpatialRGPT~\citep{Cheng2024SpatialRGPTGS} starts with facts from templates about spatial relations and uses a language model from the Llama family to create prompts. SpaRE~\citep{Ogezi2025SpaREES} begins with image captions and uses a Qwen language model to generate pairs of questions and answers, considering only those image captions that contain explicit information about a form of spatial reasoning.

LLM-annotation is useful for generating explanations, as well as step-by-step reasoning. 3D ReasonSeg~\cite{Ning2025EnhancingSR} considers the step-by-step reasoning generated by LLMs to be an explicit annotation, generating question-reasoning-answer triples all grounded in 3D spatial relations. SSR~\cite{Liu2025SSRED} instead uses the LLM-generated rationales as intermediate supervision, translating depth-derived spatial hints into a structured trace of reasoning that can be distilled into latent embeddings.
In addition, LLMs can serve as automatic annotators and reviewers in the loop. In SpatialThinker~\citep{Batra2025SpatialThinkerR3} Claude Sonnet4 is used to generate question-answer pairs and to assign a difficulty rating to each question. Afterwards, GPT-4o is used for consistency checking between questions, answers and their spatial information. This is done with the goal of preserving well-grounded and reliable samples through multi-model consistency checking.
A problem with using LLM-based annotation is the risk of hallucination or misinterpretation of the annotated visual content~\cite{Liu2024MitigatingHI}. To reduce this issue, alternative datasets utilize procedural generation and 3D reconstruction techniques to create and annotate spatial relationships directly from geometry rather than language.

\subsubsection{Procedural Generation and 3D Reconstruction}
Procedural generation allows spatial reasoning datasets to have ground-truth knowledge for concepts where it is impractical to have humans label reliably at scale. The most impressive application of this is SpatialVLM's~\citep{Chen2024SpatialVLMEV} 3D lifting pipeline, reconstructing estimated metric-scale point clouds from internet images using off-the-shelf depth and segmentation models. The reconstruction allows metric distance questions like \enquote{How far is object A from B in meters?} to be answered, which would otherwise be difficult for a human to do. RoboPoint~\citep{Yuan2024RoboPointAV} combines procedural generation with an affordance prediction setting, producing spatial training data for robots from a 3D scene plus camera pose from a simulation, yet generalizes to real robots. RoboSpatial~\citep{Song2025RoboSpatialTS} extracts robot-centric spatial annotations from real images and 3D scans, capturing reference frames and relations neglected by typical image captions. While these ideas depend on reconstructed or simulated 3D information to obtain spatial labels, SpatialMosaic~\citep{Lee2025SpatialMosaicAM} relies on it to capture a property of the view from high-quality ScanNet++ and Waymo Open Dataset point clouds. Using these point clouds, it then procedurally computes occlusion of objects across views and chooses pairs of views where the object is visible in the answer but occluded in the query. Based on these view pairs, it creates questions involving partial visibility in both indoor and outdoor environments.
In summary, these methods automatically generate spatial annotations from reconstructed or simulated 3D data. However, they can only generate annotations for properties that can be computed from the available geometry. Errors in the estimated 3D data can be propagated into the generated annotations. For desired supervision that cannot be derived from the 3D data, other annotation methods can be used in conjunction with these approaches.

\subsubsection{Model-Generated Supervision}
The annotation schemes described in Table~\ref{tab:spatial_reasoning} fix questions, answers, or spatial annotations prior to training. In addition, the methods in this section also use model-generated outputs to create preference data before optimization, or to acquire reward signals during training. These approaches can utilize the annotations from the table but do not require annotated chains of reasoning for each example. Correctness signals on answers still need either reference answers or a defined verifier.
SpatialThinker~\citep{Batra2025SpatialThinkerR3} incorporates scene graph creation into its reasoning output and then employs online reinforcement learning using rewards for answer correctness, number of regions, output formatting, and spatial relations grounding. SpaceR~\citep{Ouyang2025SpaceRRM} employs spatially guided reinforcement learning for video reasoning tasks and uses task-specific verifiable rewards as well as a mapping reward that compares the predicted locations of objects to corresponding reference locations in a grid map. Embodied-R~\citep{Zhao2025EmbodiedRCF} is applied to embodied tasks and provides rewards for correctness, logical consistency, and output formatting to encourage the reasoning to justify the final answer.
Model-generated outputs can also be used as supervision via preference learning. Fine-Grained Preference Optimization (fDPO)~\citep{Shen2025FineGrainedPO} employs Multi-Model Monte Carlo Tree Search (M3CTS) to generate reasoning traces for supervised fine-tuning and preference pairs before performing preference optimization on offline-generated responses.

\subsubsection{Human Supervision}
Human supervision provides an intermediate level of automation, which incorporates a human touch at strategic points in the pipeline. Humans can ask insightful and well-justified spatial questions, but the millions of QA pairs used to train a VLM cannot be easily annotated manually. Some hybrid models therefore use human labor in earlier stages to guide the design or the generation process. Robo2VLM~\citep{Chen2025Robo2VLMVQ}, for example, had an expert review a small set of generated question-answer pairs before full dataset generation, using the feedback to refine the prompt and generation
pipeline. Spatial-DISE~\citep{Huang2025SpatialDISEAU} takes a more design-oriented approach, manually defining a cognitively grounded taxonomy of spatial reasoning tasks before deploying an automated pipeline to scale it up. Several other works build on existing resources that are annotated by human experts. For example, SpatialThinker~\citep{Batra2025SpatialThinkerR3} draws its spatial reasoning questions from the annotated ground truth of Visual Genome. Finally, when the goal is reliable evaluation, some introduce small, manually verified benchmarks along with their training data; RoboSpatial follows this approach with a compact hand-annotated evaluation set \citep{Song2025RoboSpatialTS}.

\subsection{Data Modalities}
\label{sec:modalities}
Many datasets share the same baseline: a sample of RGB images, questions, and answers. Various datasets include additional elements beyond this baseline: some elements are derived from the visual modality (e.g., bounding boxes or masks), which do not introduce a new sensor signal but allow for the division of an image into objects. Others provide geometry or sensing channels (for example, depth maps or point clouds), which can come from sensors, reconstruction, or estimation from RGB. Other additional elements are a temporal dimension to the visual data (for example, video sequences or trajectories of objects within those videos).  Table~\ref{tab:spatial_reasoning} lists the data modalities of each dataset.

\paragraph*{Vision and Language.}
The simplest form of these datasets involves pairing the RGB images with language questions. VLMs are fine-tuned in these settings to act as conditional text generators, and VQA is a model-agnostic label type. A few datasets do, however, exclude vision altogether and instead ask the models to perform spatial reasoning based on language alone~\citep{Li2024ReframingSR}. Such settings help to determine whether the failure of VLMs to answer spatial language questions is due to failures in perception (vision) or reasoning (language) alone.

\paragraph*{Bounding Boxes and Segmentation Masks.} 
Although bounding boxes, segmentation masks, and keypoints are often treated as inputs, they are a form of structure that can be provided by human annotators or computed from RGB images or geometry. These features are used to break apart the visual scene into separate, named objects. For instance, the language model might be asked to recognize that the mug is to the left of the laptop, which requires the model to understand the locations of both objects in the visual scene. Furthermore, these features are often an output of the models rather than an input: models such as RoboPoint~\citep{Yuan2024RoboPointAV} and \citet{Ranasinghe2024LearningTL} explicitly encode these coordinates as text tokens to force the models to ground their language in the visual scene, while models such as SpatialThinker~\citep{Batra2025SpatialThinkerR3} use these coordinates as a target output to ensure the models can recognize the objects they are asked to reason about. Finally, SpatialLadder~\citep{Li2025SpatialLadderPT} learns to recognize these objects as part of the first stage of its curriculum to enable later spatial reasoning abilities.

\paragraph*{Depth Maps and Point Clouds.}
While supervision from RGB-only images does not uniquely offer information regarding the metric distances within the image, which is essentially the lack of capability of VLMs~\citep{Cai2025SpatialBotPS,Chen2024SpatialVLMEV}, there are different ways of incorporating 3D information into the models. The geometry information may be incorporated either during the generation of the answer to the question, or it may be provided as an input to the model itself. For instance, SpatialVLM~\citep{Chen2024SpatialVLMEV} utilizes monocular RGB images to reconstruct a point cloud through depth and camera pose estimation methods, using that 3D information to generate questions and answers that are used to supervise the learning of the model; the model is provided only with the RGB image and the question to be answered during inference. 
SpatialLadder~\citep{Li2025SpatialLadderPT} utilizes the same approach for their models, but instead uses existing 3D reconstructions from ScanNet to obtain the 3D information. In InternSpatial~\citep{Deng2025InternSpatialAC}, both methods are used based on the dataset used. For 3D datasets, existing geometry and camera information are used, while for image-only datasets, depth and camera information are estimated.

Other approaches, such as SpatialBot~\citep{Cai2025SpatialBotPS}, provide both RGB and depth data from an image (in the form of RGB-D images) to the model and use the depth data in answering questions during inference. Depth-trained models achieve over 99\% performance on the depth task in SpatialBench compared to approximately 84--90\% for the RGB-only inputs, and about twice the end-to-end performance in pick-and-place experiment~\citep{Cai2025SpatialBotPS}. However, this is not always the case. Using the same depth image to provide to GPT-4o in zero-shot results in a drop from 70.6 to 61.8 in the accuracy of the object's position. Thus, the benefits of depth information depend on if the model is trained to utilize that information. The distinction also has implications for deployment cost. Running a model with depth added during dataset construction, adds no depth processing cost at inference time, while the model-time depth requires the presence of a depth sensor to gather information or an estimator and then processing it whenever the model is running. The answer to whether the runtime input for depth sensors is worth the cost appears to be somewhat dependent on the domain in which the models are deployed. Two datasets that have compared 2D versus 3D models indicate different results. RoboSpatial~\citep{Song2025RoboSpatialTS} created a dataset that is ready for both 2D and 3D models and trains both types of models, showing a benefit to training with 3D models for robotics applications. Open3D-VQA~\citep{Zhang2025Open3DVQAAB} also created a 3D dataset but for aerial applications, where their 3D-LLM had better scores for estimating distance. But this distance advantage was not consistent across the other tested 3D models over 2D ones. SSR~\citep{Liu2025SSRED} takes a middle path of estimating depth from a monocular RGB image and compressing that depth into latent tokens for the model, allowing the model to benefit from depth information without requiring a depth sensor at inference time.

\paragraph*{Video and Trajectories.}
While static images can express spatial relations like \enquote{above} or \enquote{left of}, modeling dynamic spatial relations like \enquote{moving into} requires some notion of motion within the ground truth. Thus, there is a second main family of spatial datasets that was built with the goal of incorporating video supervision and trajectories. Such videos can enable new capabilities beyond static images alone, such as understanding the appearance of objects in a sequence of images (InternSpatial~\citep{Deng2025InternSpatialAC}), computing displacement of objects or tracking motion in videos (Video-STR~\citep{Wang2025VideoSTRRM}, and tracking objects that appear in various viewpoints (VILASR~\citep{Wu2025ReinforcingSR}). There are, actually, two different senses of the concept of \enquote{dynamic} relations. One sense is that the model’s input is a video itself (SpaceR~\citep{Ouyang2025SpaceRRM}, Video-STR~\citep{Wang2025VideoSTRRM}, Embodied-R~\citep{Zhao2025EmbodiedRCF}). The other is that the model’s reasoning is to be dynamic to the static image - for instance, the model may be provided with a static image but instead be asked what the scene will look like after it has been rotated (Spatial-DISE~\citep{Huang2025SpatialDISEAU}, Open3D-VQA~\citep{Zhang2025Open3DVQAAB}). Thus, these two different types of dynamic relation enable the model to provide answers about the motion it observed and the hypothetical transformations.

\subsection{Source Environment}
The environments that are represented in these spatial reasoning datasets closely reflect the target uses. Datasets intended for robotics prioritize indoor and tabletop environments, reflecting their intended tasks~\citep{Song2025RoboSpatialTS,Chen2025Robo2VLMVQ}. RoboPoint also targets indoor environments, but does so almost exclusively using procedurally generated simulated environments, rather than real-world captures~\citep{Yuan2024RoboPointAV}. 
SpatialMosaic \citep{Lee2025SpatialMosaicAM} covers both types of environments. It uses indoor scans from ScanNet++ and outdoor driving sequences from the Waymo Open Dataset specifically in order to avoid over-specialization to indoor environments. The dataset also provides both indoor and outdoor evaluations separately.
General-purpose spatial reasoning datasets, on the other hand, sample from a wider variety of everyday environments, including indoor and outdoor scenes with people, objects, and activities~\citep{Deng2025InternSpatialAC,Cheng2024SpatialRGPTGS}. Recent work expands on these environments by SpaceVista, featuring indoor, outdoor, and aerial (drone-view) video domains from a variety of scales (mm to km) in order to enable spatial reasoning across many physical scales~\citep{Sun2025SpaceVistaAV}.


\section{Evaluation}
\label{sec:evaluation}
After training an MLLM on spatial reasoning, it is necessary to assess whether the targeted spatial reasoning skill improved and that the general vision-language abilities of the model have not been harmed during the training process. While most surveyed works in the field report on both aspects of the evaluation, a notable minority of studies do not report on the retention of general vision-language capability in addition to the improvement in the targeted capability. Table~\ref{tab:eval-practices} summarizes the existing evaluation techniques for fine-tuning foundation models.

\paragraph*{What is measured.}
The evaluation of the VLMs targeted at spatial reasoning skills uses a variety of measures. While benchmarks report one overall score for spatial accuracy, they also provide separate scores for different categories of spatial reasoning skills such as perception, relations, and metric reasoning. SPAR-Bench, used alongside the SPAR-7M training dataset, outputs scores for depth and distance estimation, position matching, view-change inference, object relations, and spatial imagination~\cite{Zhang2025FromFT}. Spatial-DISE~\citep{Huang2025SpatialDISEAU} separates spatial tasks into two cognitive axes: intrinsic (the internal parts and structure of a single object) vs. extrinsic (relations among several objects or to a frame) and cross-evaluates them with static vs. dynamic relations. Spatial-DISE finds that their models fall below the human baseline in different categories, including intrinsic-dynamic spatial relations, such as mentally rotating or folding one object. These failures correspond to the specific types of spatial estimations encoded in their datasets and could be averaged away by reporting a single number. Furthermore, the fine-tuned models are also evaluated on general vision-language benchmarks: SpaRE~\citep{Ogezi2025SpaREES} on MMMU, MMBench, and HallusionBench tests, and fDPO~\citep{Shen2025FineGrainedPO} on MME and POPE tests. While post-training helps improve performance on spatial tasks, performance on general tasks is not consistently retained in the reported results. SPAR's score on MME decreases from 2186 to 2163 over its matched control and from 2323 to 2163 over the untrained base~\citep{Zhang2025FromFT}. For RoboSpatial, the MME-cognition score drops from 305.4 to 293.2, but all other general scores remain the same or increase~\citep{Song2025RoboSpatialTS}. Additionally, the trend of decreased performance across studies is difficult to follow due to variations in baselines. For example, fDPO's general benchmark is evaluated relative to SpatialRGPT instead of its initial model and therefore does not indicate retention in its general capability~\citep{Shen2025FineGrainedPO}. Spatial-DISE~\citep{Huang2025SpatialDISEAU} has also found that fine-tuning on only portions of spatial data exhibits benefits only on external tasks that are of the same type (quadrant-specific experiments). For instance, if a model is only fine-tuned on viewpoint transformation then transfer to other less related tasks is \enquote{largely minimal or even detrimental}. However, when the model is fine-tuned on the entire of the DISE dataset then performance is improved on all external tasks. These findings imply that transfer is dependent on both the type of training performed and the type of evaluation tasks, and not that task similarity should be the sole explanation for transfer. General-benchmark results can only indicate retention relative to an appropriate initial checkpoint using the same evaluation protocol. While comparisons to other models can provide information about performance, they do not indicate what was gained or lost by the adapted model.

\paragraph*{How it is scored.}
The choice of metric is contingent upon the type of answer that the dataset asks. For datasets that require categorical answers (left/right, yes/no, or multiple choice), accuracy and exact match are used to score the models~\citep{Zhang2025FromFT,Huang2025SpatialDISEAU}. For answers that are provided as a metric (distance traveled, area of interest, etc.), exact match is often not a meaningful evaluation metric, so methods that consider whether the models' answers fall within a certain tolerance of the ground truth values are used for evaluation; these can be metrics like $\pm$25\% relative-tolerance success rates~\citep{Cheng2024SpatialRGPTGS}, or mean relative accuracy~\citep{Ouyang2025SpaceRRM,Wu2025ReinforcingSR}. For datasets that ask for open-ended answers, the answers from the models are provided to a language model to determine whether they are correct~\citep{Liu2025SSRED,Deng2025InternSpatialAC}. LLMs can also assist in judgment or numerical extraction alongside tolerance-based scoring~\citep{Cheng2024SpatialRGPTGS,Shen2025FineGrainedPO,Zhang2025Open3DVQAAB}.For answers that ask for actions to be performed by a robot (points to be selected, trajectories to follow, etc.), there are two general approaches: some datasets use an algorithm to determine whether the predicted action is geometrically correct. For example, in RoboSpatial, a predicted 3D point is considered correct if it lies within the convex hull of a set of reference points computed from the scene geometry; the authors describe this as deliberately conservative~\citep{Song2025RoboSpatialTS}. The second approach is to evaluate whether the model succeeded with a downstream task of interest performed by the robot; i.e., RoboPoint~\citep{Yuan2024RoboPointAV} and SpatialBot~\citep{Cai2025SpatialBotPS} ask whether the robot can complete a specific physical action after receiving an answer from the model. Finally, many of these datasets, such as Robo2VLM~\citep{Chen2025Robo2VLMVQ} and Spatial-DISE~\citep{Huang2025SpatialDISEAU}, also provide human baselines to compare against human competence, not only against each other.

\paragraph*{How generalization is tested.}
Many recent evaluations have sought to separate in-domain from out-of-domain evaluations to determine whether the models are learning a generalizable skill or merely adapting to the structure of one particular benchmark, since a model can score well by memorizing benchmark-specific patterns like answer formats or phrasing without actually learning to reason spatially, and out-of-domain tests can reduce reliance on those familiar cues, making it harder to succeed without genuine generalization. SpatialLadder~\citep{Li2025SpatialLadderPT} reports gains on benchmarks beyond their own splits, including CV-Bench, SPAR-Bench, and ViewSpatial-Bench. While these benchmarks vary in distribution and format, improvements across them do indicate some level of cross-benchmark transfer. However, they are not sufficient to completely eliminate the possibility of shared annotation practices, biases across tasks, or common sources of data. VILASR~\citep{Wu2025ReinforcingSR} evaluates on a variety of different task categories, spanning different input types and spatial reasoning demands, to demonstrate that the model is not reliant upon a particular regime of inputs or spatial condition. Furthermore, many of these models have evaluated their approach on real-world scenes to assess whether they can generalize their knowledge from being trained on synthetic scenes to the real world~\citep{Zhang2025Open3DVQAAB,Yuan2024RoboPointAV}. Finally, additional evaluations beyond accuracy can help to gauge generalizability. These range from lower-level checks, like testing sensitivity to wording by evaluating on paraphrased questions~\citep{Cheng2024SpatialRGPTGS}, to broader ones like checking whether learning transfers to held-out environments~\citep{Song2025RoboSpatialTS}, or not using training videos that overlap with the evaluation set to avoid measuring transfer across already-seen environments~\citep{Ouyang2025SpaceRRM}.

\begin{table}[hbt]
    \centering
    \caption{
    Evaluation practices across the training datasets of Table~\ref{tab:spatial_reasoning}.
    Codes are defined in the legend below.
    }
    \label{tab:eval-practices}
    \footnotesize
    \setlength{\tabcolsep}{4.5pt}
    \renewcommand{\arraystretch}{1.02}
    \begin{tabular}{llcccc}
        \toprule
        \textbf{Citation} & \textbf{Answer type / scoring} & \textbf{Retent.} & \textbf{Independ.} & \textbf{Robust.} & \textbf{Human} \\
        \midrule
        \cite{Song2025RoboSpatialTS}                    & cat, act / acc, geom        & \checkmark & \checkmark & env   & --         \\
        \cite{Chen2025Robo2VLMVQ}                       & cat / acc                   & --         & --         & --    & \checkmark \\
        \cite{Ning2025EnhancingSR}                      & cat, seg / acc, iou         & -- & --         & --    & --         \\
        \cite{Liu2025SSRED}                             & cat, met, open / acc, judge & \checkmark & --         & --    & --         \\
        \cite{Deng2025InternSpatialAC}                  & cat, met, open / acc, judge & \checkmark & --         & --    & \checkmark         \\
        \cite{Ranasinghe2024LearningTL}                 & cat, act / acc, geom        & \checkmark & \checkmark & --    & --         \\
        \cite{Zhang2025Open3DVQAAB}                     & cat, met, open / acc, tol   & --         & --         & s2r   & --         \\
        \cite{Yuan2024RoboPointAV}                      & act / geom, task            & \checkmark & \checkmark & s2r   & --         \\
        \cite{Ogezi2025SpaREES}                         & cat / acc                   & \checkmark & \checkmark & --    & --         \\
        \cite{Tang2025SparkleMB}                        & cat, met / acc              & \checkmark & --         & --    & --         \\
        \cite{Huang2025SpatialDISEAU}                   & cat / acc                   & -- & --         & --    & \checkmark \\
        \cite{Ouyang2025SpaceRRM}                       & cat, met / acc, mra         & \checkmark & \checkmark & mod   & --         \\
        \cite{Batra2025SpatialThinkerR3}                & cat / acc                   & \checkmark & --         & --    & --         \\
        \cite{Shen2025FineGrainedPO}                    & cat, met / acc, judge, tol              & -- & --         & --    & --         \\
        \cite{Wu2025ReinforcingSR}                      & cat, met / acc, mra         & --         & \checkmark & --    & --         \\
        \cite{Sarch2025GroundedRL}                      & cat / acc                   & \checkmark & \checkmark & --    & --         \\
        \cite{Chen2025PerceptionBR}                     & cat / acc                   & \checkmark         & \checkmark & --    & --         \\
        \cite{Zhao2025EmbodiedRCF}                      & cat / acc                   & \checkmark         & --         & --    & --         \\
        \cite{Chen2025ThinkW3}                          & cat, met / acc, mra                   & \checkmark & --         & --    & --         \\
        \cite{Cai2025SpatialBotPS}                      & cat, met, act / acc, task   & \checkmark & --         & --    & --         \\
        \cite{Li2025SpatialLadderPT}                    & cat, met / acc, mra         & \checkmark         & --         & --    & --         \\
        \cite{Cheng2024SpatialRGPTGS}                   & cat, met / acc, tol         & \checkmark & --         & para  & --         \\
        \cite{Chen2024SpatialVLMEV}                     & cat, met / acc, tol         & \checkmark & --         & --    & -- \\
        \cite{Wang2025VideoSTRRM}                       & cat, met / acc, mra         & \checkmark & \checkmark & --    & --         \\
        \cite{Sun2025SpaceVistaAV}                      & cat, met / acc, mra         & --         & --         & mod    & \checkmark         \\
        \cite{Zhang2025FromFT}                          & cat, met / acc, mra              & \checkmark & --         & --    & \checkmark         \\
        \cite{Lee2025SpatialMosaicAM}                     & cat, met / acc, tol                   & --         & --         & env    & \checkmark         \\
        \cite{Cai2025ScalingSI} & cat, met, open / acc, judge & \checkmark & \checkmark & dbias & --      \\
        \midrule
        \multicolumn{2}{l}{\emph{Count}} & 20/28 & 10/28 & 8/28 & 6/28 \\
        \bottomrule
    \end{tabular}

    \vspace{4pt}
    \begin{minipage}{\textwidth}
    \footnotesize
    \setlength{\parindent}{0pt}

    \textbf{Answer type.} Categorical (\textit{cat}), metric (\textit{met}), open-ended (\textit{open}), action or point (\textit{act}), segmentation (\textit{seg}). A dash (--) denotes not reported, here and in every column.

    \textbf{Scoring.} Accuracy or exact match (\textit{acc}), relative tolerance band such as $\pm25\%$ of the ground-truth value (\textit{tol}), mean relative accuracy (\textit{mra}), LLM-as-judge for semantic correctness (not only to extract an answer) (\textit{judge}), geometric criterion such as convex-hull containment (\textit{geom}), intersection-over-union (\textit{iou}), downstream task success (\textit{task}).

    \textbf{Retent.} General-capability retention reported on non-spatial vision--language benchmarks with matched control. A checkmark denotes an assessment, not the absence of decline in performance.

    \textbf{Independ.} The primary evaluation set is independent of the authors' own training-data construction pipeline. Several entries also serve as benchmarks, for which independence is assessed against the accompanying training split.

    \textbf{Robust.} Probe beyond headline accuracy: paraphrased questions (\textit{para}), held-out environments (\textit{env}), modality shift (\textit{mod}), sim-to-real transfer (\textit{s2r}), debiasing or shortcut probes such as no-vision and circular testing (\textit{dbias}).

    \textbf{Human.} Human task performance baseline reported. This is distinct from human-annotated ground truth, data verification, rating model outputs, and baselines from existing benchmarks.

    \textbf{Count.} Final row; counts are over the 28 works listed.

    \end{minipage}
\end{table}


\section{Insights and Future Directions}

\label{sec:findings_and_future_directions}

\subsection{Cross-Dataset Observations}

When the datasets are considered collectively rather than separately, there are several patterns that do not emerge so readily from individual studies. For example, in some training phases, perceptual grounding of objects in an image appears to precede spatial reasoning about those same objects. This is not just a matter of style, but also yields better results in certain experiments. Studies such as SpatialLadder~\citep{Li2025SpatialLadderPT} and \citet{Chen2025PerceptionBR} explicitly incorporate perceptual tasks (spatial awareness) before their reasoning-oriented training phases (such as spatial cognition). SpatialThinker~\citep{Batra2025SpatialThinkerR3} and Sparkle~\citep{Tang2025SparkleMB} provide evidence for the advantages of having foundational spatial awareness and rewards before more advanced training, although they do explore varying training orders. The convergence of these studies across different methods and systems implies that there is, in fact, value in having foundational spatial awareness, but they do not prove that a strict training order is required.

Relatedly, staging of the training of these models can lead to improved results at the corpus sizes of most of these datasets. Essentially, staging controls the order in which skills are introduced and built upon; for instance, training can be staged to introduce perceptual tasks first, supervised spatial reasoning second, and reinforcement learning third, so that each stage builds on the skills established by the one before it. This type of staging is implemented within the two-stage variant of \citet{Chen2025PerceptionBR}, which trains perception before reasoning within reinforcement learning itself.
SpatialLadder~\citep{Li2025SpatialLadderPT}, also contains three stages and moves from object localization to multi-dimensional spatial understanding to reinforcement learning with verifiable rewards. They report 43.9 on VSI-Bench for perception-then-spatial training, compared with 42.7 for spatial-only training and 40.7 for mixed training.

Another trend involves the balance between scale and supervision design. There have been successes with large corpora of synthetic data with broad coverage~\citep{Chen2024SpatialVLMEV,Deng2025InternSpatialAC}; however, SpatialThinker demonstrates that a significantly smaller corpus can also be effective when combined with verifiable spatial rewards. SpatialThinker is trained on \(\sim 7{,}000\) QA samples and performs better than both supervised fine-tuning and conventional reinforcement-learning methods on fourteen tasks~\citep{Batra2025SpatialThinkerR3}. While this benchmarking does not imply that rewards can always replace the need for large training data (since models and datasets differ), it does imply that alongside pretrained knowledge, carefully designed verifiable rewards can lessen the reliance on extremely large static training corpora.

The impact of reinforcement learning also varies based on when it is applied in the training pipeline. SpaceR benchmarks reward-based training against supervised fine-tuning from the same initial point and observes that reward-based training has higher overall performance across all tasks tested~\citep{Ouyang2025SpaceRRM}. However, in ViGoRL, it is found that applying GRPO to the base model results in less spatially grounded reasoning compared to the warm-start method~\citep{Sarch2025GroundedRL}. Thus, reinforcement learning should not be thought of as uniformly better or substituting for supervised learning. Instead, comparisons should only be made while holding constant the previous level of supervision and the point in the training pipeline at which reinforcement learning is applied.

As discussed in Section~\ref{sec:modalities}, depth and geometry are valuable as a supervision signal, while their value as runtime inputs relies on the task and the model. In datasets with large amounts of data, depth is often used during data collection and then training occurs using only RGB~\citep{Chen2024SpatialVLMEV,Deng2025InternSpatialAC}, which provides effective supervision.  At inference time, providing geometry to the model is less clear, and the available evidence across domains is mixed: In robotics, geometry seems to provide benefits~\citep{Song2025RoboSpatialTS}, while in aerial images, Open3D-VQA~\citep{Zhang2025Open3DVQAAB} reports a distance-estimation advantage for 3D-LLM, while fine-tuned Qwen2-VL-7B has the higher overall score.

Finally, reference frames and viewpoints appear to be a common source of challenge, and their related terminology differs across datasets. Several datasets that test multiple frames or viewpoints consistently report that models struggle with frame transformations and perspective-taking, including RoboSpatial~\citep{Song2025RoboSpatialTS} across ego, world, and object frames, Open3D-VQA~\citep{Zhang2025Open3DVQAAB} on allocentric-egocentric transformation, and InternSpatial~\citep{Deng2025InternSpatialAC} on multi-view reasoning. At the same time, terms such as \enquote{multi-view}, \enquote{egocentric}, and \enquote{object-centric} are used inconsistently across papers.

\subsection{Evaluation Open Problems}
While there has been a lot of work done, there still lacks a unified way to evaluate spatial reasoning and having shared evaluation metrics across works in Table~\ref{tab:eval-practices}. According to our observations, there are four issues that make it difficult to compare across works and reach conclusions about the efficacy of specific methods. Firstly, some works utilize LLMs to evaluate open-ended spatial answers~\citep{Liu2025SSRED,Deng2025InternSpatialAC}. However, the LLMs used might make mistakes in spatial interpretation or answer understanding. Additionally, agreement between the LLM evaluations and human spatial evaluations is not reported in these works. SSR compares a second judge, GPT-4o-mini, with its original evaluator. But this comparison does not imply agreement with human evaluations~\citep{Liu2025SSRED}.

Secondly, the threshold for what constitutes a \enquote{tolerated} error in the evaluation tasks is often defined separately in each paper. SpatialRGPT uses a relative error tolerance of $\pm$25\% of the ground-truth value~\citep{Cheng2024SpatialRGPTGS}, and VILASR uses mean relative accuracy, which computes accuracy across ten relative accuracy thresholds from 0.50 to 0.95~\citep{Wu2025ReinforcingSR}. Because threshold values vary across papers, a model can exhibit higher accuracy simply by using a looser threshold, which makes accuracy values across different benchmarks not directly comparable. Thirdly, for some spatial benchmarks both the training and test sets are created using the same construction process. Because of this, improvements in accuracy observed between two benchmarks created in this way could be partly due to similarities in the construction process. Lastly, transfer changes with the training configuration and the spatial tasks used for evaluation. This means that high performance on benchmarks could indicate only narrow performance on specific tasks rather than general spatial reasoning ability. Overall, evaluating accuracy on a single benchmark provides an incomplete view of a model’s spatial reasoning ability.

\subsection{Future Directions}
Building on these observations, we outline several directions for future work:

\paragraph*{Unified dynamic annotation.}
In Section~\ref{sec:modalities} we discussed how motion datasets are split between those that require dynamic input (video) and those that only require dynamic reasoning across static images. Furthermore, motion, trajectory, visibility and affordance annotations are generally done via different and task-specific annotation systems~\citep{Wang2025VideoSTRRM,Yuan2024RoboPointAV,Lee2025SpatialMosaicAM,Zhao2025EmbodiedRCF}. There is a need for a more unified system of dynamic annotation that can simultaneously capture spatial, temporal, and affordance information, as well as the physics that link these to space.

\paragraph*{Efficient reasoning.} 
Reinforcement learning can improve multi-step spatial reasoning, but SpaceR shows that it can also produce long reasoning chains~\citep{Ouyang2025SpaceRRM}. These chains are expensive to decode, and high-resolution images also increase visual-token costs~\citep{Yang2025VisionThinkSA}. Future datasets could include reasoning chains of different lengths, allowing models to reason more only when needed.

\paragraph*{Toward more comparable evaluation.}  
In Section~\ref{sec:evaluation} existing evaluation methods are quite fragmented. Different works use different scoring methods, have different tolerance levels, use different benchmark datasets, and use LLM evaluators whose agreement with human spatial judgments is not reported. The improvements that are observed sometimes only transfer to similar tasks. A unified evaluation scheme using similar tasks, agreed tolerance ranges, common baselines, and clearly reported scoring procedures would enable more direct comparisons across datasets and would also help guide future research.

\paragraph*{Spatial scale and geospatial data.} Most of the datasets discussed in this survey operate at what Montello~\citep{Montello1993ScaleAM} refers to as figural and vista scale; that is, they contain information about small objects, tabletops, and rooms. While there are a few exceptions to this (such as SpaceVista’s ability to cover scales from mm to km~\citep{Sun2025SpaceVistaAV}, or Open3D-VQA’s aerial image understanding~\citep{Zhang2025Open3DVQAAB}), there are also a few separate datasets dedicated to geospatial tasks. For instance, there are datasets based upon remote sensing~\citep{Kuckreja2024GeoChatGL,Li2024VRSBenchAV} or geospatial chain-of-thought questions~\citep{Liu2025TowardsFR}, though these are generally separate from the types of datasets surveyed in this paper. How spatial reasoning can be transferred from one of these categories to the other, however, is an area for future research.

\paragraph*{Human-centred settings.} 
Spatial understanding is important for safe human-robot interaction~\citep{Sisbot2007SpatialRF}, yet the datasets we surveyed seldom capture its interactive dimension: a robot reasoning about space jointly with a person, where human intention and bidirectional interaction shape the spatial context. Datasets designed around such interaction would address a dimension that current embodied datasets largely leave open.


\section{Conclusion}
\label{sec:conclusion}
This survey examined spatial reasoning in fine-tuned MLLMs and VLMs from a data-centric point of view, with a special focus on various approaches used in the literature for annotation, design, and dataset building. We analyzed datasets used for post-training and classified the literature based on our taxonomy into three distinct categories: dataset design, building, and evaluation. Moreover, we organized them based on modality, annotation approach, framing, and view. Spatial reasoning ground-truth signals, perception and reasoning, and also framing in 3D space were comprehensively discussed from a data perspective. At the core of the paper, we carefully studied different dataset-building strategies and approaches, which give readers unique insights into vital design decisions before building datasets, given their applications and practical constraints in building large-scale datasets. After performing cross-dataset comparisons, we found several trends. Perception in some training curricula is established prior to higher-level spatial reasoning. There are also reports of benefit from staged training. Verifiable rewards may help reduce dependency on large fixed training sets in certain approaches. Depth and geometry are often helpful in creating supervision but show mixed results as runtime inputs. We conclude the survey by highlighting open problems and inconsistencies in evaluation protocols and possible future directions in spatial reasoning dataset development. These findings suggest that future progress depends not only on larger datasets, but also on more consistent task definitions, reference frames, supervision strategies, and evaluation protocols.

\section*{Statements and Declarations}

\subsection*{Funding}

We acknowledge the financial support of the Pan-Canadian AI Strategy, led by CIFAR, and the Alberta Machine Intelligence Institute (Amii).

\subsection*{Author contributions}

Mahdieh Mallahnezhad: Conceptualization; Methodology; Investigation; Formal analysis; Writing--original draft; Writing--review \& editing; Project administration. Masoud Jafaripour: Conceptualization; Investigation; Writing--original draft; Visualization; Writing--review \& editing. Mohamadali Tofigh: Investigation; Formal analysis; Visualization; Writing--review \& editing. Sheila Schoepp: Conceptualization; Investigation; Formal analysis; Visualization; Writing--review \& editing. Osmar R. Zaiane: Conceptualization; Supervision; Writing--review \& editing.


\bibliography{sn-bibliography}

\end{document}